\documentclass[letterpaper, 10 pt, conference]{ieeeconf}  

\IEEEoverridecommandlockouts                              

\usepackage{mathrsfs}
\usepackage{graphicx}
\usepackage{stfloats}
\usepackage{amsmath}
\usepackage{cases} 
\usepackage{overpic}
\usepackage{color}
\usepackage[T1]{fontenc}
\usepackage[skip=1ex]{caption}
\usepackage{booktabs,tabularx}
\usepackage{siunitx}
\newcolumntype{C}{>{\centering\arraybackslash}X}

\title{\LARGE \bf
	Fast-Learning Grasping and Pre-Grasping\\ via Clutter Quantization and Q-map Masking
}

\author{Dafa Ren$^{1}$, Xiaoqiang Ren$^{1}$, Xiaofan Wang$^{1}$, S. Tejaswi Digumarti$^{2}$ and Guodong Shi$^{2}$
	\thanks{$1$: School of Mechatronic Engineering and Automation, Shanghai University, Shanghai, 200444, China. {\tt\small E-mails: (dafaren, xqren, xfwang)@shu.edu.cn}}
	
	\thanks{$2$:  Sydney Institute for Robotics and Intelligent Systems,  The University of Sydney, Sydney, NSW 2006, Australia. {\tt\small E-mails: (tejaswi.digumarti, guodong.shi)@sydney.edu.au}}%
}%

\begin{document}

\maketitle
\thispagestyle{empty}
\pagestyle{empty}

\begin{abstract}
	
	Grasping objects in cluttered scenarios is a challenging task in robotics. Performing pre-grasp actions such as pushing and shifting to scatter objects is a way to reduce clutter. 
	Based on deep reinforcement learning, we propose a Fast-Learning Grasping (FLG) framework,  that can integrate pre-grasping actions along with grasping to pick up objects from cluttered scenarios with reduced real-world training time. We associate rewards for performing moving actions with the change of environmental clutter and utilize a hybrid triggering method, leading to data-efficient learning and synergy. Then we use the output of an extended fully convolutional network as the value function of each pixel point of the workspace and establish an accurate estimation of the grasp probability for each action. We  also introduce a mask function as prior knowledge to enable the agents to focus on the accurate pose adjustment to improve the effectiveness of collecting training data and, hence, to learn efficiently. We carry out pre-training of the FLG over simulated environment, and then  the learnt model is transferred   to the real world with minimal fine-tuning for further learning during actions. 
	Experimental results demonstrate  a 94$\%$ grasp success rate and  the ability   to generalize to novel objects. Compared to  state-of-the-art approaches in the literature, the proposed FLG framework can achieve similar or higher grasp success rate with lesser amount of  training in the real world. Supplementary video is available at https://youtu.be/e04uDLsxfDg.
	
\end{abstract}

\section{INTRODUCTION}

Grasping objects is one of the basic tasks of robotic manipulation. 
While it may seem trivial for a human to pick an object from a pile, it is quite challenging to train a robotic manipulator to do the same.
One of the first challenges is due to the presence of clutter in the environment.
Partial visibility of an object of interest and obstacles obstructing the manipulator, in an object-rich environment, impede a successful grasp.
Secondly, in a typical industrial or logistics automation application, objects to be grasped or avoided are constantly updated.
This requires the development of algorithms that can generalize well to previously unseen objects.
Furthermore, collecting data from real robots to train a grasping algorithm is expensive and time consuming.
Hence, an ideal robotic grasping solution should be able to tackle these challenges.

\begin{figure}
	\centering
	\includegraphics[width=1\columnwidth]{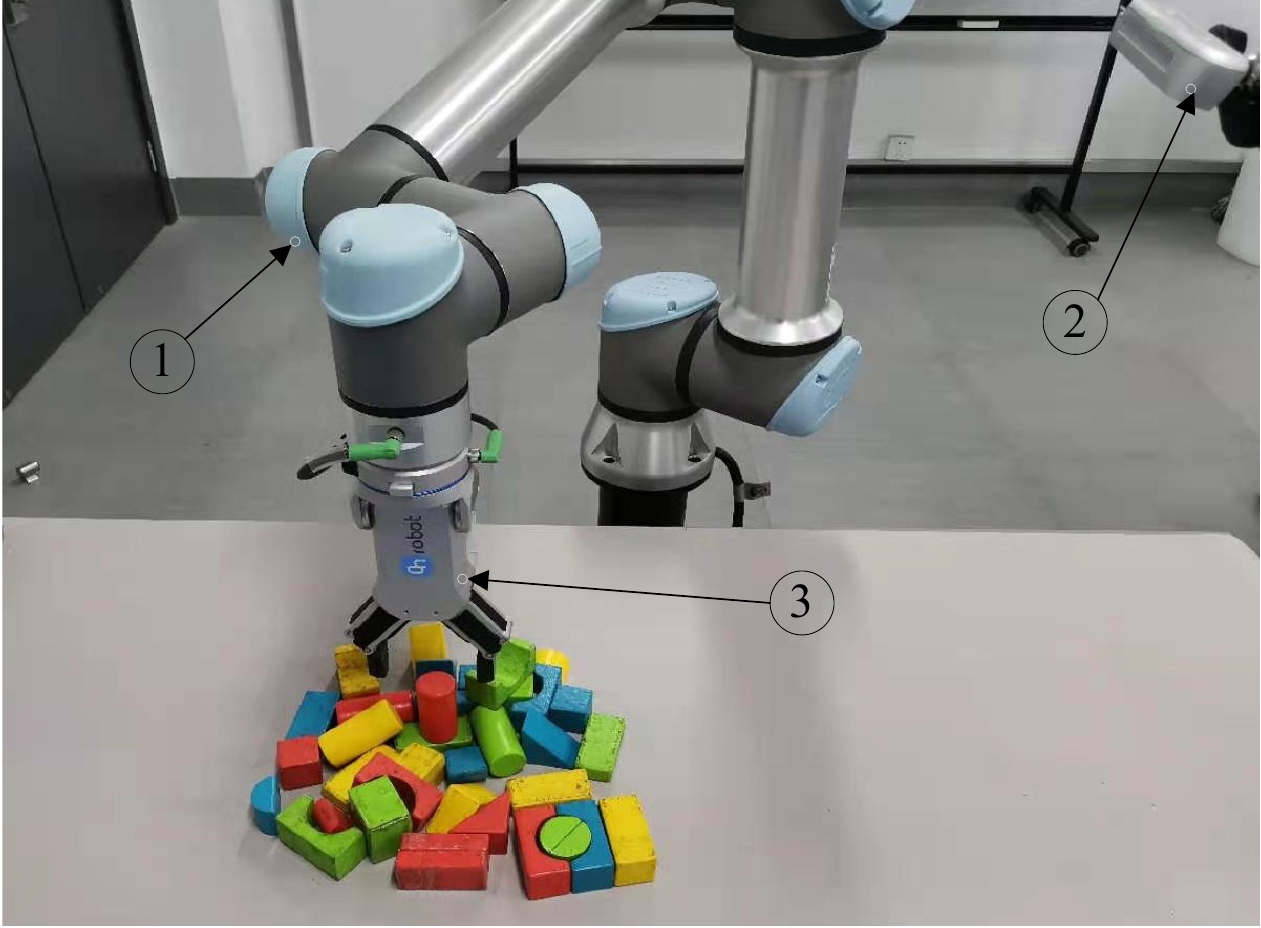}
	\caption{Our robot is able to grasp objects in highly cluttered environments. Our hardware setup consists of (1) a UR5e robot arm, (2) an Intel Realsense RGB-D camera and (3) an RG2 gripper.}
	\label{framework}
\end{figure}

One way to address clutter in the environment is to perform a pre-grasping actions, such as moving or shifting objects around, to facilitate grasping. 
In recent years, such approaches are being actively studied~\cite{zeng2018learning, 8968042} and are gaining popularity.
However, the approaches developed so far either have a low success rate or need a long training time in real world, which is not amenable to the short deployment times desired by the industry.

In this paper, we propose a learning based framework (see Fig.~\ref{framework}) that enables accurate grasping in cluttered environments with minimal real-world data.
To achieve this, we first define two pre-grasping manipulation primitives: pushing and shifting. We use these primitives to increase the probability of grasping.
We then train an end-to-end self-supervised learning network without human interaction based on model-free deep reinforcement learning in simulation.
This network, when transferred to real world, performs strongly and generalizes well to novel objects.
To be specific, the key aspects of our system are:

\begin{itemize}
	\item We introduce clutter quantization maps to characterize environmental clutter. Using these maps, we design a novel pre-grasping reward function that associates rewards for moving actions with the change of environmental clutter. 
	\item We propose a shifting action, in addition to pushing, which helps in dealing with clutter better.
	\item We introduce a mask function which promotes efficient learning
	by focusing on meaningful and precise pose adjustment. 
	\item We extend the fully convolutional network (FCN) for reward estimation. The network trained in the simulation environment is able to reach a success rate of more than~90$\%$ only in 2.5 robot hours in the real-world training.
	
\end{itemize}

\section{RELATED WORK}

Robotic manipulation, and in particular grasping, are active areas of research within robotics. Classic analytical approaches~\cite{bohg2013data} find stable force-closure for known objects by utilizing 3D models of objects and their physical properties. Recent years have witnessed a spurt of progress in deep learning based computer vision and thus data-driven approaches to robotic manipulation have become a research hotspot.

Data-driven approaches can be divided into two groups based on whether they are model-based or not. Model-based data-driven approaches usually sample grasps using the combination of object detection~\cite{shi2019pointrcnn}, pose estimation~\cite{kehl2017ssd, li2019cdpn, li2018deepim} and grasp estimation~\cite{20196, liang2019pointnetgpd} to pick the objects. But these approaches usually leverage object specific knowledge (i.e., shape, pose). Consequently their ability to generalize to unknown object classes and novel shapes is limited. While real-world datasets with a large amount of labelled data, such as~\cite{jiang2011efficient, depierre2018jacquard, fang2020graspnet}, have been developed to overcome this issue, real-world data collection and labelling is still time consuming, laborious and expensive. 

In contrast, some data-driven approaches explore model-agnostic grasping strategies that directly link visual data to candidate grasps~\cite{zeng2018robotic, 7487517, kumra2020antipodal}. Our approach is based on model-free deep reinforcement learning. Using reinforcement learning enables us to learn more extensive grasp representation through exploitation and exploration in a self-supervised way. Model-agnostic learning also shows great results for generalizing to novel objects~\cite{8968042}.

Handling clutter is another active area of research. To mitigate collisions introduced by clutter, pre-grasping actions such as shifting and pushing, have been proposed as manipulation primitives~\cite{yang2020deep, deng2019deep, RN1}. 
Kalashnikov \textit{et al.}~\cite{kalashnikov2018qt} trained a QT-opt network for grasping, resulting in a grasping success rate of 96$\%$ for generalizing to unknown objects. In addition, the robots implicitly learned other non-prehensile actions like pushing. However, their robot setup relied upon over 580K real-world grasp attempts on 7 real robotic systems. Berscheid \textit{et al.}~\cite{8968042} used shifting actions to increase the grasping probability and removed the need of shifting sparse rewards by making shifting directly dependent on the change in the success rate of grasping. Their system improved the grasp rate to 98.4$\%$, at the cost of around~25000 grasp and 2500 shift training data. Both of these methods consume a large number of real-world training data. In this work, we train out network in a simulation environment and then fine-tune the pretrained network in the real world.

More closely related to our work is that of Zeng \textit{et al.}~\cite{zeng2018learning}, which introduced a $Q$-learning framework to learn the complementary pushing and grasping strategies simultaneously. They utilized a Fully Convolutional Network (FCN) as a function approximator to estimate the $Q$ function of the reinforcement learning framework. This approach used significantly less training data than~\cite{kalashnikov2018qt}, but the success rate was lower. Analogous to this method, our pre-grasping actions include not only a pushing primitive, but also an additional shifting primitive to further enlarge grasping task scenarios. We also extend the FCN by utilizing a combination of skip connections \cite{peng2019pvnet} and upsampling to improve learning efficiency. Our experiments demonstrate that our method is capable of grasping objects more accurately.

\section{APPROACH}

\begin{figure}
	\centering
	\includegraphics[width=1\columnwidth]{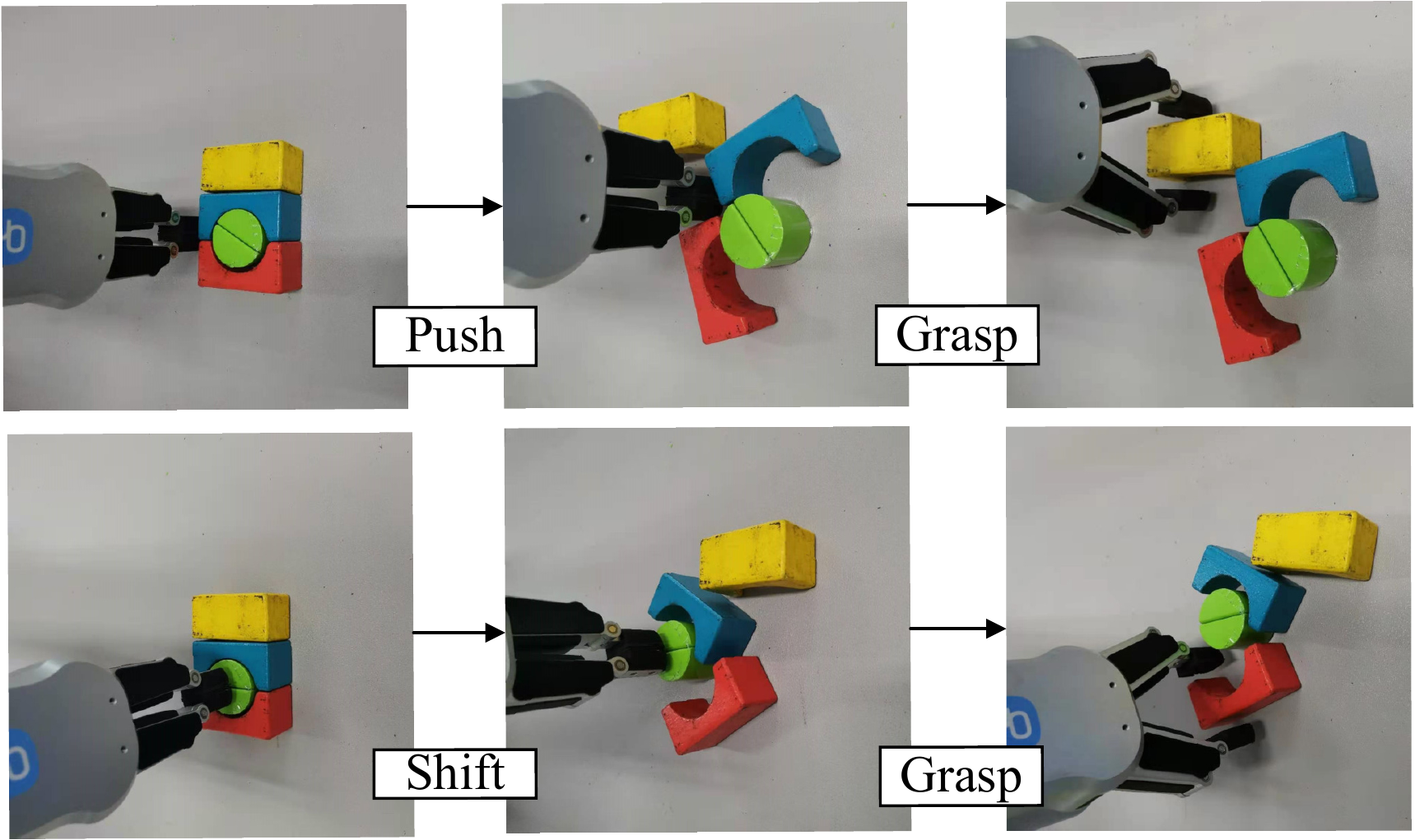}	
	\caption{\textbf{Manipulation primitives.} When faced with a scene where the target cannot be directly grasped, the robot can first scatter objects by pre-grasping actions such as push and shift, and then grasp them.}
	\label{Manipulation primitives}
\end{figure}

\begin{figure*}
	\centering
	\begin{overpic}[width=1\textwidth]{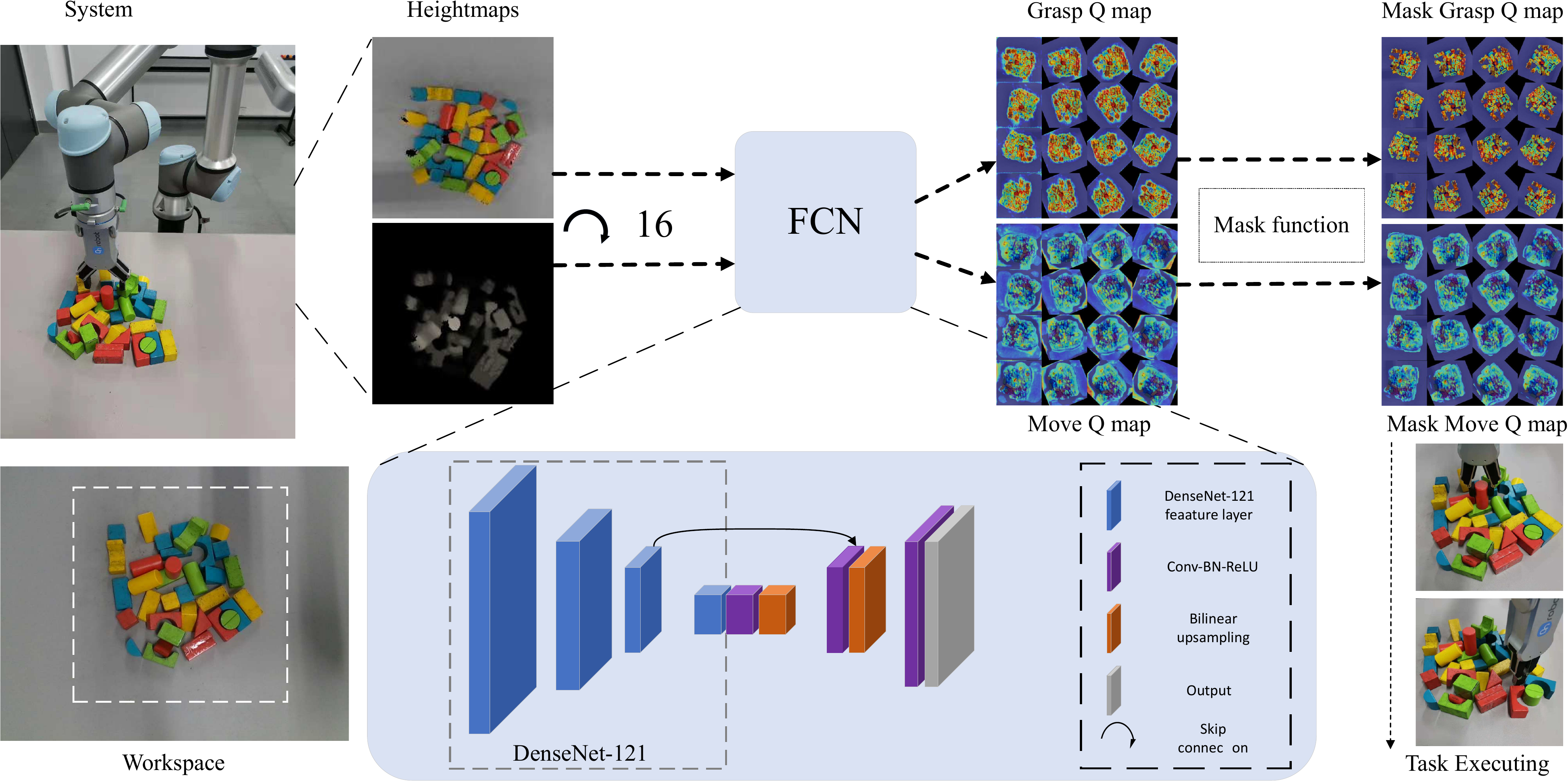}
	\put(39,35){$\times$}
	\end{overpic}
	\caption{\textbf{Overview.} The visual 3D data observed by a statically mounted RGB-D camera is orthographically projected to construct heightmaps. Then, the heightmaps are rotated by 16 orientations for angles $\theta_i=\frac{2\pi i}{16}$ and then fed into the FCN. With the combination of skip connection and upsampling, the FCN model outputs dense pixel-wise maps of Q values. Each pixel of dense pixel-wise map predicted by FCN represents a different location on which to execute the primitive in the corresponding angle. Finally, our robot arm performs the primitive selected by mask function.}
	\label{overview}
\end{figure*}

We formulate the task of robotic grasping as a Markov decision process (MDP), which is defined by its tuple $(\mathcal{S}, \mathcal{A}, r)$ with the state space $\mathcal{S}$, the action space $\mathcal{A}$, and the reward \mbox{function $r$}. We train the the robot using deep reinforcement learning (Q-learning), whose goal is to find the solution to the MDP, which is a policy $\pi : \mathcal{S}\rightarrow \mathcal{A}$ mapping the current state $s_t \in \mathcal{S}$ to an action at $a_t \in \mathcal{A}$ that maximizes the sum of expected rewards. Considering the huge action space and associated sparse rewards, we limit our process to a discrete action space to improve sample efficiency.

\subsection{Manipulation primitives} We parameterize the action space to a tuple $(x, y, z, \theta_i, \phi_j)$, where $(x, y)$ is given by dense pixel-wise map of $Q$ values outputted by the deep network, $z$ is the height of the predicted point, $\theta$ is the rotation angle around the $z$-axis and $i$ corresponds to the index of the rotation angle with a size of 16 in our system (i.e., $\theta_i = \frac{2\pi i}{16}$). The last parameter $\phi$ corresponds to the top-down grasping action ($\phi_g$) or the preset moving action ($\phi_m$). The moving action can either be a shifting or a pushing action.

In the execution phase of the action, the end effector of the robotic manipulator reaches the $(x, y)$ coordinate given in the prediction map with $z$ height, at an angle $\theta_i$. When the controller selects $\phi_g$, the end effector grasps from top to bottom; when the controller selects $\phi_m$, the moving action is a shifting or pushing action which is determined by the height of the predicted point, which indicates whether there is a target there. If there is a target object at the predicted location, the end effector will choose the shifting action, otherwise a pushing action is taken to disperse the targets which cannot be grasped due to obstruction (see Fig.~\ref{Manipulation primitives}).

\subsection{Rewards}
We design reward functions for grasping and moving actions as follows. 
When the end effector performs the grasping action and reaches the target position, the end effector closes, and then the robot returns to a pre-specified \emph{home} position. If the antipodal distance between the fingers of the gripper is greater than zero upon reaching the \emph{home} position, it is a successful grasp. This corresponding reward $r_g$ for this grasping action is formulated as

\begin{equation}
	r_g (s) =
	\begin{cases} 
		1,  & \mbox{if successful grasping,} \\
		0, & \mbox{else.}
	\end{cases}
\end{equation}

The purpose of the moving action is to increase the probability of objects that can be successfully grasped, in an environment with dense clutter or ungraspable objects. This is done by scattering the objects. The moving reward should therefore be related to the change in the environmental clutter. Notice that a potentially good moving action will result in an increase of coverage or a change of target position. For this reason, we parameterize the change of environmental clutter by the change of the heightmap value $\mu(t)$ and the change of coverage value $\eta(t)$. The heightmap change function $\mu(t)$ is formulated as

\begin{equation}
	b_i(t) =
	\begin{cases} 
		1,  & \mbox{if $a_i(t+1)-a_i(t)>\delta$,} \\
		0, & \mbox{else.}
	\end{cases}	
\end{equation}
\begin{equation}
	\mu (t) = \sum_{i=1}^N b_i
\end{equation}
where $a_i(t)$ is the height above the work surface at pixel $i$ at time $t$, $\delta$ is a preset threshold, and $N$ is the total number of pixels in the heightmap (here, 224 $\times$ 224). The height $a_i(t)$ is computed by projecting depth values, acquired from an overhead RGB-D camera, orthographically onto the work surface, i.e. the table.

\begin{figure}
	\centering
	\includegraphics[width=1\columnwidth]{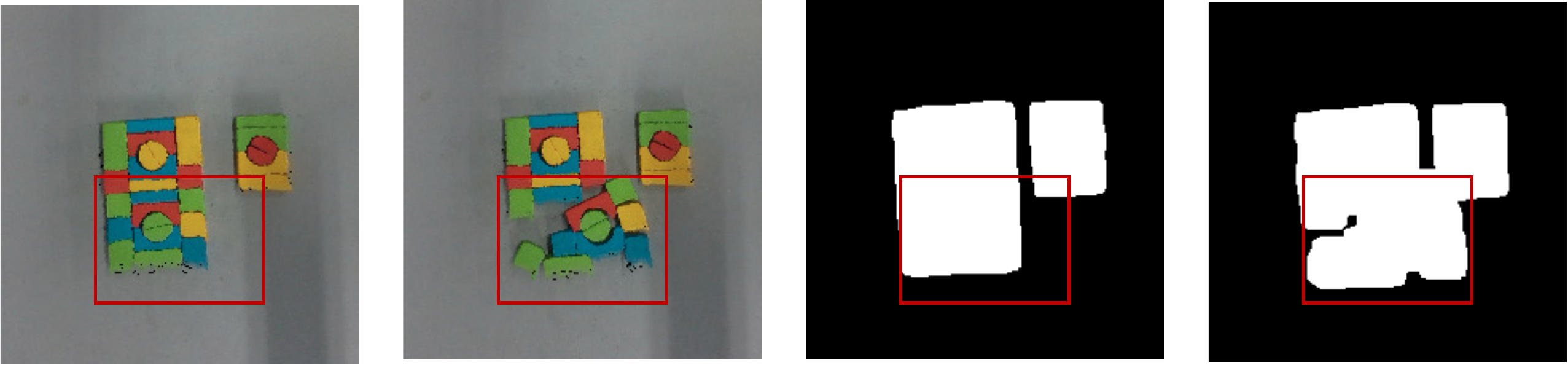}
	
	\caption{\textbf{Clutter quantization map.} A potentially good moving action will increase the number of pixels with a value of 1 (white pixels) in the clutter quantization map (right) corresponding to the color image (left).}
	\label{mask}
\end{figure}

To define $\eta(t)$, we introduce \emph{Clutter Quantization Map} (see Fig.~\ref{mask}), which is the heightmap after binarization and dilation. Specifically, each pixel of heightmap is first set as 1 if the height value is larger than the minimum height of the workspace and 0 otherwise. Then, a dilation is applied to set to 1 the neighbours of each pixel that has a value of 1. The change in coverage $\eta(t)$ is formulated as

\begin{equation}
	\eta(t) = \sum_{i=1}^N x_i(t+1) - \sum_{i=1}^N x_i(t)
\end{equation}
where $x_i(t)$ are the pixel value of the clutter quantization map at time $t$, and $N$ is the number of pixels in the heightmap. Notice that the dilation step fills gaps between adjacent objects and that objects that are scattered have increased area as opposed to when they were clustered together. Thus, this reward encourages scattering.

We now define a novel moving reward function as
\begin{equation}
	r_m (s) =
	\begin{cases} 
		0.5,  & \mbox{if $\mu>\tau_1$ or $\eta > \tau_2$,} \\
		0, & \mbox{else.}
	\end{cases}
\end{equation}
where both of $\tau_1$ and $\tau_2$ are preset thresholds.

\subsection{Network}

In designing the network architecture, we draw inspiration from the framework proposed by Zeng \textit{et al.}~\cite{zeng2018learning}. However, there are several key differences that help our proposed framework outperform theirs. 

We employ a dueling-DDQN~\cite{wang2016dueling} as the core of our framework. Unlike normal DDQN~\cite{van2016deep}, which can only update the value of one action per iteration, the value stream $V$ of dueling-DDQN is updated with every update of the $Q$ values. In addition, in our picking task, multiple actions corresponding to multiple targets may share comparative rewards for a given state. Furthermore, the robustness of the updates is also improved. 

We use a Fully Convolutional Network (FCN) based on DenseNet-121~\cite{8099726} to model the $Q$ function of our dueling-DDQN (see Fig.~\ref{overview}). Each heightmap and corresponding color image is rotated 16 times by an angle of $\pi/8$ radians, which together form the state inputs. The network predicts dense pixel-wise maps of $Q$ values of the same size as the output. Every pixel in the $Q$ value map represents a grasping or moving action primitive at the corresponding 3D location. The rotation angle of the heightmap corresponds to the rotation angle of the end effector.  The dense pixel-wise parameterization greatly simplifies the action space and speeds up the convergence speed.

The feature layer output of DenseNet-121 is bilinearly upsampled to~4 times after 1 $\times$ 1 convolutional layer interleaved with ReLU and batch normalization, followed by channel-wise concatenation with the corresponding intermediate feature of DenseNet-121. The resulting concatenated feature then passes through Conv, ReLU, Batch normalization layers and additional 4 times bilinear upsampling to obtain a high-precision dual-channel dense pixel-wise map, each of which corresponds to grasping map and moving one, respectively. 

Unlike in~\cite{zeng2018learning}, where two separate FCNs are trained for the grasp and push actions, we train a single FCN which outputs a dual-channel pixel wise map for each of the actions. In our experiments, our simple yet efficient network structure takes only~0.7s to calculate the next action even if the size of action space \mbox{is $224 \times 224 \times 32$}, which is much faster than Zeng \textit{et al.}~\cite{zeng2018learning} (i.e., 1.2s). In addition, the combination of skip connections~\cite{peng2019pvnet} and bilinear upsampling in our FCN further boosts accuracy.

\subsection{Exploration Policy}

\begin{figure}
	\centering
	\includegraphics[width=1\columnwidth]{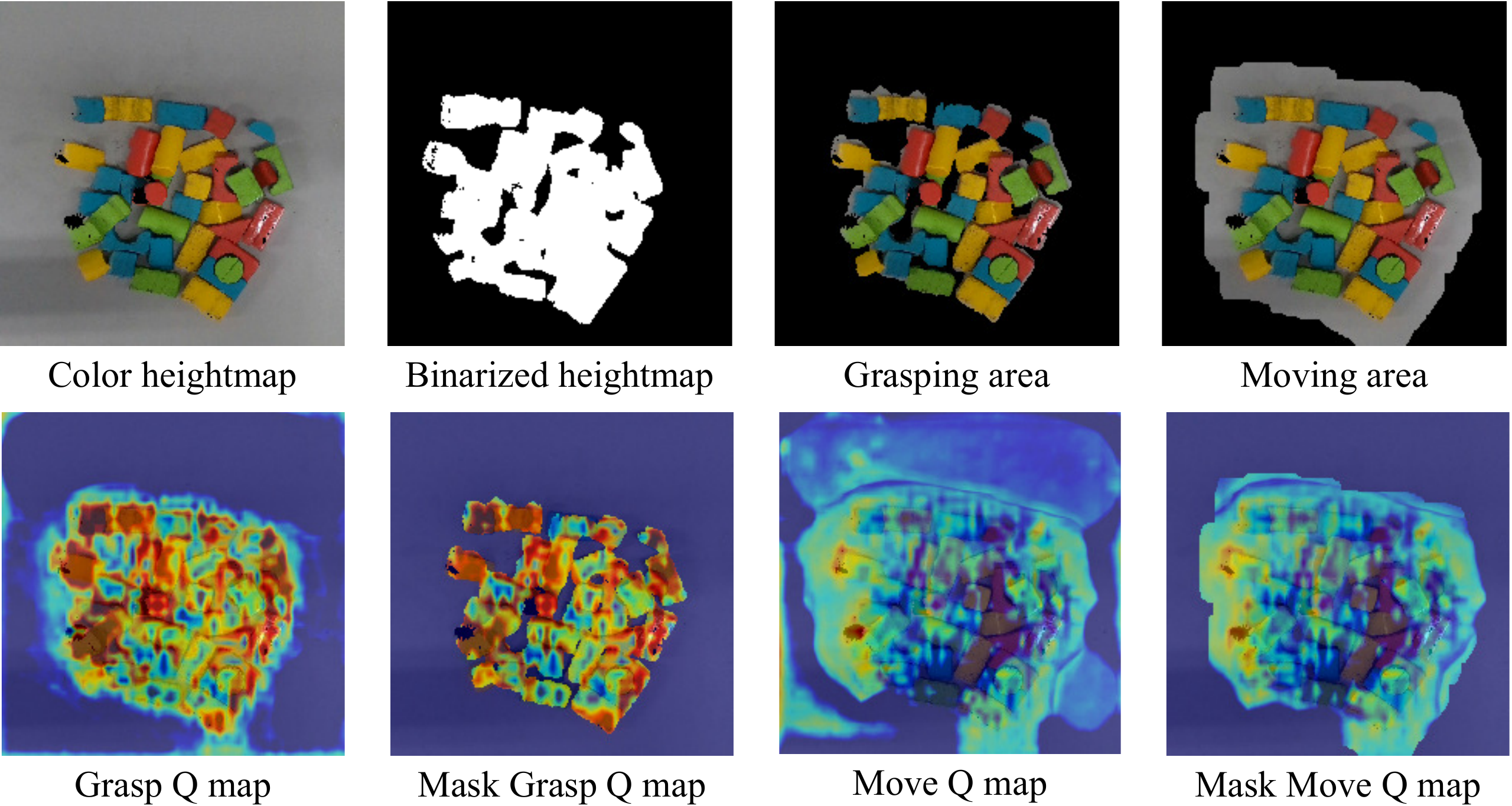}
	
	\caption{\textbf{Mask function.} The system performs the grasping or moving action in the area (top row) with high success probability compulsively after using mask function. The dense pixel-wise maps of $Q$ values are visualized as heat maps (bottom row).}
	\label{motion}
\end{figure}

Since efficient exploration contributes to data-efficient learning, we introduce two masks namely the grasping mask $\rho_g$ and the moving mask $\rho_m$ and incorporate them into the exploration policy. The system makes an action decision based on the prior probability maps
\begin{equation}
	\pi_e: \mathop{\arg\max}_{a}  \rho \circ Q
\end{equation}
where $\circ$ is the Hadamard product, $\rho$ may be $\rho_g$ or $\rho_m$ according to the action taken, and~$Q$ are the dense pixel-wise maps of $Q$ values. We filter the $Q$ maps by the masks, the effect of which is illustrated in the bottom row of Fig.~\ref{motion}.

\noindent$\displaystyle \textbf{Grasping Mask $\rho_g$}$ is a binary mask, generated from the heightmap by setting a pixel to 1 if the height value at that pixel is greater than the minimum height of the workspace and 0 otherwise. Since the model parameters are initialized randomly, in the early stages of training, the network will most likely predict incorrect grasping positions and try to grasp in the object-less background area. As a result, the training data essentially consists of negative samples. The mask $\rho_g$ prevents the robot from persisting exploration in these object-less areas and thereby encourages the system to focus on the adjustment of accurate grasping pose in the desired target areas (see Fig.~\ref{motion}).

\noindent$\displaystyle \textbf{Moving Mask $\rho_m$}$ is generated from $\rho_g$ by dilation, because the areas of interest for a moving action are the target objects and their surrounding areas. The training samples for moving actions occur relatively less frequently than those of grasping actions as they appear only in scenarios where a grasp is not possible. Hence, using this mask helps the system to focus solely on the areas of interest and thus provide more meaningful training data.

To improve the exploratory degree and sample diversity, our exploration policy employs $\epsilon$-greedy strategy. Since the system iteration has one-step lag and the last action may not change the state, the system may make the same wrong decision as the previous step, which reduces the validity of data and training speed. To cope with this, if the execution location of the current step is the same as the previous step, the action in the previous step will not be chosen at the current step. We will select one of the actions corresponding to the $M$ maximum action-value function uniformly (here, $M=10$). Through this strategy, our system avoids potential repeated failures.

\section{EXPERIMENTAL RESULTS}

The system uses an AMD 3970X processor and an NVIDIA RTX 2080Ti for computing. Our system adopts prioritized experience replay~\cite{schaul2015prioritized}, the loss function uses the Huber loss function, and the optimizer is the Adam optimizer.

We design a series of experiments to evaluate the proposed approach, which we call \textbf{Fast-Learning Grasping (FLG)}. The goals of the experiments are 1) to demonstrate that our approach can accurately grasp objects in different levels of cluttered scenarios, 2) to show the significance of the novel network structure and the exploration policy, 3) to test whether our algorithm is capable of learning grasping effectively on a real system, and 4) to investigate whether our model can directly generalize to novel objects. 

\subsection{Simulation Experiments}

\begin{figure}
	\centering
	\includegraphics[width=0.8\columnwidth]{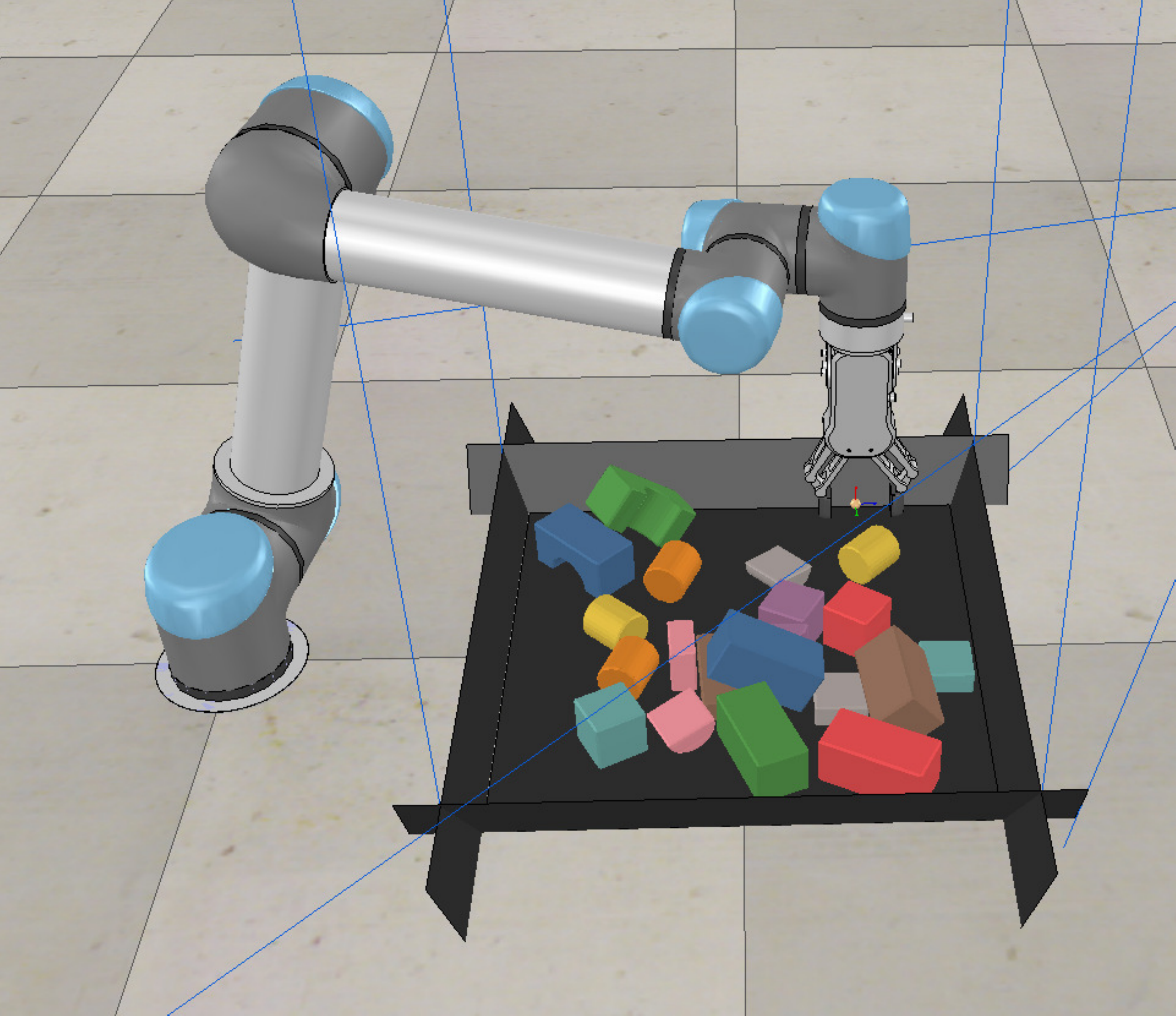}
	
	\caption{\textbf{Simulation environment.} Policies are trained in scenarios with random arrangements of 20 objects.}
	\label{sim}
\end{figure}

\begin{figure}
	\centering
	\includegraphics[width=1\columnwidth]{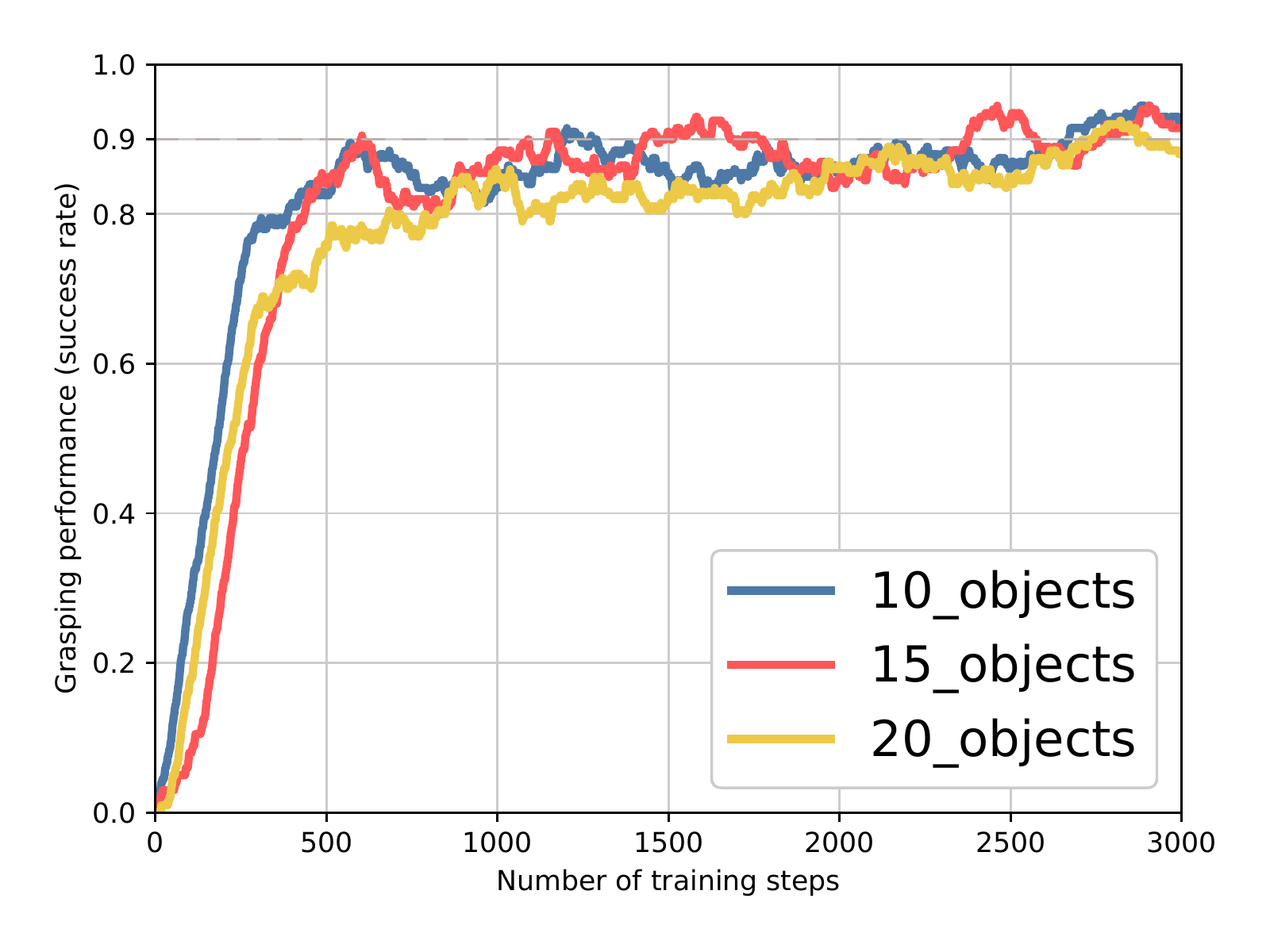}	
	\caption{Comparison of performance of our policies trained with different number of objects. The success rate of each step is the average success rate of the first 200 steps of the current step.}
	\label{dif-obj}
\end{figure}

Our simulation environment consists of a UR5 robot arm with an RG2 gripper in CoppeliaSim~\cite{zeng2018learning} (shown in~Fig.~\ref{sim}). We compare performance of our policies trained with different number of objects (see Fig.~\ref{dif-obj}). When we train the model in simulation with 10 objects, the system achieves a grasping success rate of 80$\%$ for 400 action attempts and a grasping success rate of 90$\%$ for 550 action attempts. In the same setting, the method of Zeng \textit{et al.}~\cite{zeng2018learning} requires more than a thousand training steps to achieve a success rate of about 80$\%$. Our system outperforms their method with regard to training speed and success rate. Although the training time is slightly extended in the more cluttered environment, the final success rate is higher than~90$\%$. It can be seen in Fig.~\ref{dif-obj} that our model can accurately grasp objects in different levels of cluttered scenarios.

\begin{figure}
	\centering
	\includegraphics[width=1\columnwidth]{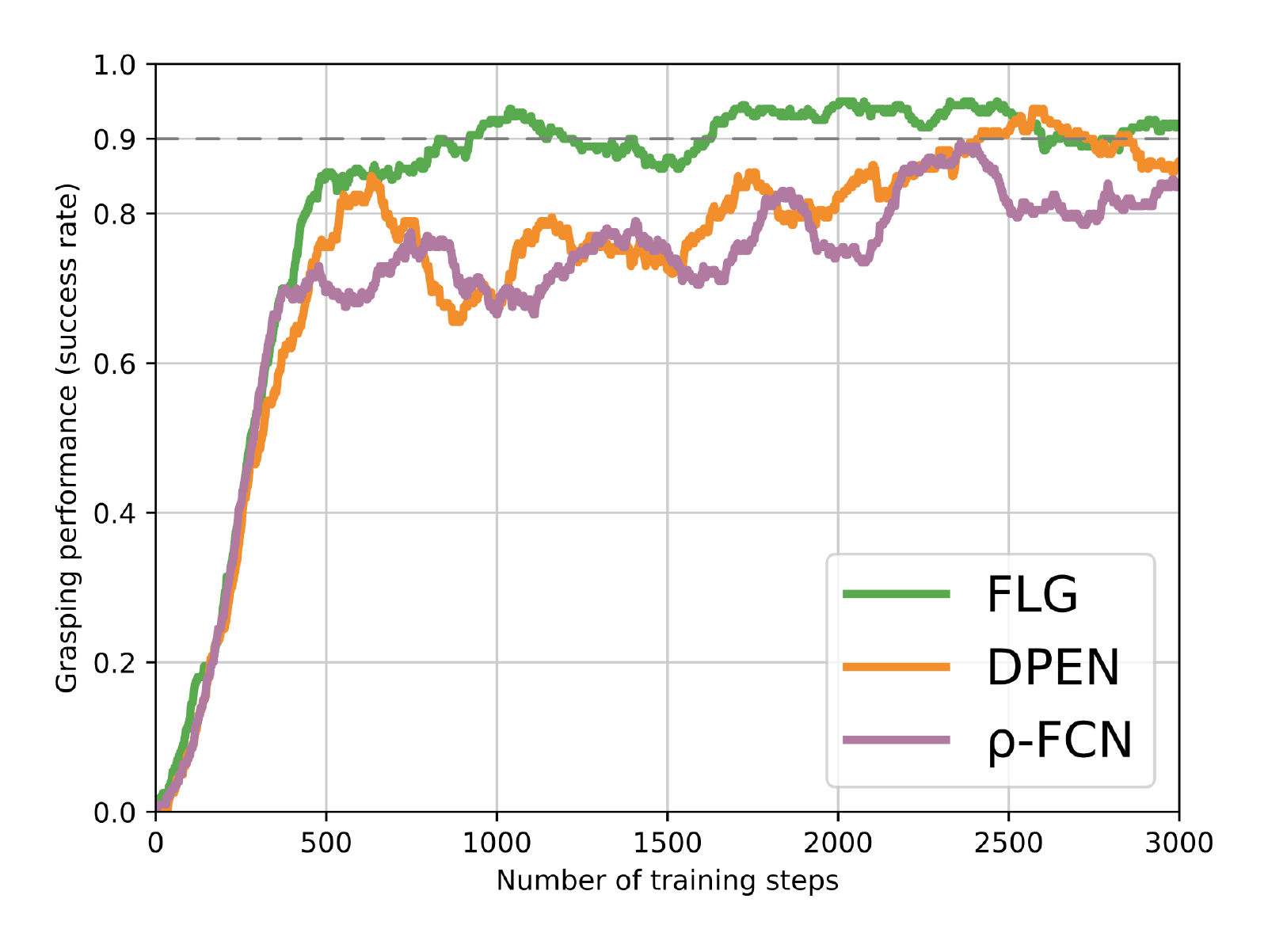}
	
	\caption{\textbf{Comparisons to Baselines.} Taking the scenario of 13 targets as an example, we compared the performance of our policy to the two baselines. Our approach (FLG) achieves a higher and more robust grasping success rate in shorter training time in comparison to baseline architectures.}
	\label{baseline}
\end{figure}

We compare the controller's pickup performance with the two baseline methods to verify the impact of the exploration policy and network structure on the training speed and grasping accuracy: 1) \textbf{Dense Pixel-wise Estimation Network (DPEN)} is the same as our method in the network structure, but without our masks. 2) \textbf{$\rho$-FCN} is an extension of VPG~\cite{zeng2018learning} by incorporating our masks. The FCN of VPG directly takes the outputs of the feature layer of DenseNet-121 as the final $Q$ maps by sampling up 16 times without skip connections. We see that our method outperforms both baseline methods with regards to training speed and success rate (see Fig.~\ref{baseline}). The poor performance of  \textbf{$\rho$-FCN} is likely due to the outputs of the network being bilinearly upsampled to 16 times. As a result, one prediction point corresponds to a patch area in workspace. The final $Q$ maps are of low precision compared to ours. Moreover, the results of our experiments show that the grasping success rates of the models without the masks fluctuate by more than 15$\%$, and the training process is relatively unstable. In particular, a relatively low success rate is maintained for a long time in the early stages of training. In summary, we infer that using masks in the model facilitates data-efficient learning while using an extended FCN in the model improves the success rate. 

Simulations are an attractive approach for training agents as they provide a good starting point and alleviate certain safety concerns during the training process for real-world applications~\cite{peng2018sim}. 
We design a specific training process to facilitate transferring of our model from simulation to the real world. 
First of all, inspired by the idea of curriculum learning~\cite{bengio2009curriculum} and domain randomization~\cite{8202133}, we randomly place 10 blocks with different shapes and colors into the workspace at the beginning of training. The system basically does not need moving action in the case of fewer objects, thus, we artificially reduce the number of moving actions at the first~500 steps. When the iteration reaches 400 steps, we then increase gradually the number of blocks until 20 are reached into the workspace. After 1500 steps, whenever the robot empties the workspace, the environment is initialized into a preset highly cluttered scene with a probability 0.2. From our observations, the chances that the robot fails to grasp twice in a row are close to zero, under normal conditions when the model learns to grasp accurately. If the robot fails to grasp objects in two consecutive attempts, we consider the environment to be highly cluttered and that a moving intervention is needed. In this case, the system triggers a moving action to scatter the objects and reduce clutter.

\subsection{Real-World Experiments}

In the real world, we also use a UR5e robot with an RG2 gripper as the end effector. RGB-D images are captured from an Intel RealSense D415 mounted rigidly above the workspace. During training, the robot automatically adds~20 objects into the workspace at random. Then, it grasps and places all of them back until the workspace is empty and then restarts. In order to better understand the ability of our model to meet the needs of different scenarios, we train two models in the real world: 1) \textbf{Grasping-only} to learn high-precision grasping strategy quickly, which makes the model focus on finely adjusting grasping posture to grasp objects that are harder to grasp. 
Its advantage is to empty the workspace quickly with high action efficiency. 2) \textbf{Grasping-Moving} strategy is learning synergies between grasping and moving, which is capable of emptying highly cluttered scenes with high grasping success rate.

\begin{figure}
	\centering
	\includegraphics[width=1\columnwidth]{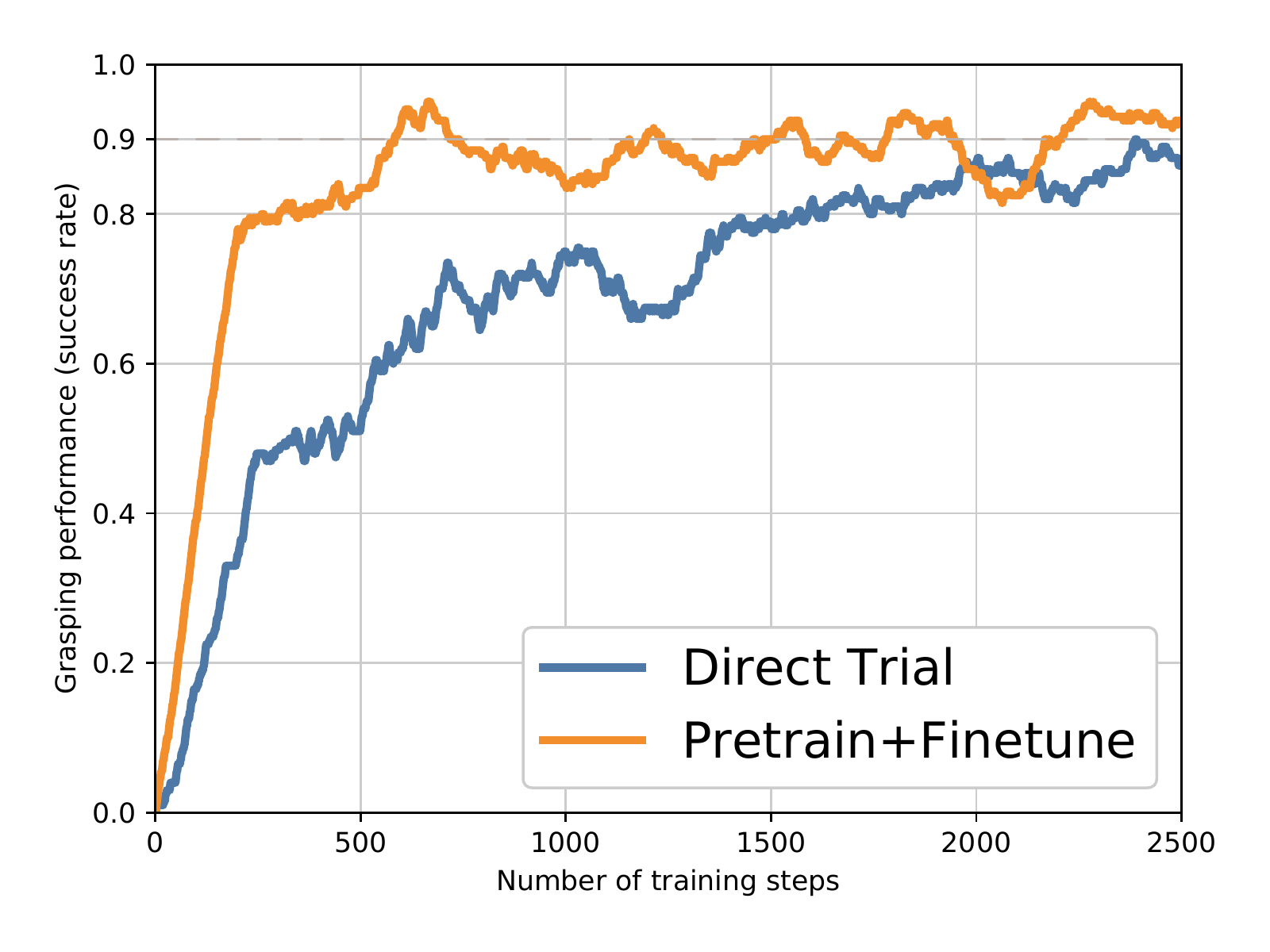}	
	\caption{Grasping performance of our algorithm with and without pretraining from the simulation environment in grasping-only mode.}
	\label{real}
\end{figure}

First of all, in the grasping-only mode, we compare the performance difference between the model directly trained only on real data and the model pretrained in the simulation environment and fine-tuned with real-world data. The training curve is shown in the Fig.~\ref{real}. It can be seen that the pretrained model has a higher performance in the early training stage, which achieves a success rate of 90$\%$ with only 700 iterations. The pretrained model has 94$\%$ success rate after fine-tuning, while the one directly trained on real data needs nearly 2500 steps to achieve 90$\%$ success rate. This experiment shows that the pretrained model with fine-tuning is more effective than the model directly trained from the real robot since the simulation data contains significantly more diverse grasping samples. 

\begin{table}[htbp]  
	\centering 
	\caption{\label{t1}Comparison of performance of state-of-the-art grasping algorithms based on Reinforcement Learning.}   
	\setlength{\tabcolsep}{0.9mm}{
		\begin{tabular}{lcccc}    
			\toprule    
			Method    & Grasp Success Rate    & Iteration Steps   \\
			\midrule
			QT-opt~\cite{kalashnikov2018qt}    & 88$\%$             & 580000                         \\
			VPG~\cite{zeng2018learning}      & 68$\%$             & 2500                               \\
			Berscheid \textit{et al.}~\cite{8968042}    & 98$\%$            & 27500                                \\
			Song \textit{et al.}~\cite{9126187}        & 92$\%$             & 15000          \\
			\textbf{Ours}                           & \textbf{94$\%$}             & \textbf{2500}                                \\    
			\bottomrule   
	\end{tabular}}  
\end{table}
The experimental data also shows that the sampling efficiency of our method is extremely impressive. The superiority of our method is evident when we compare the number of training steps required to achieve a similar success rate with other state-of-the-art methods. Although the running environment is slightly different, the results in Table~\ref{t1} illustrate the learning efficiency of our approach. While the other approaches require tens of thousands of steps when fine-tuning their models to achieve a success rate of around 90$\%$, our approach in comparison requires less than 1/10th of the number of training iterations. Even the model directly trained from the real robot performs better than the other approaches. The real robot also only needs 9s for each motion. Furthermore, our method achieves a~90$\%$ success rate, at the cost of training in the real world for only 2.5 hours. The number of picks and places is about~350 per hour. It exceeds the number reported by Berscheid \textit{et al.}~\cite{8968042}, which is 274 picks per hour.

Our algorithm requires the orthographic image of the depth image provided by RGB-D camera as input, which poses a hidden challenge. Noise in the depth images can lead to a in a decrease in the success rate of grasping, due to incorrect heightmaps. Surprisingly, during our experiments, when the depth image was corrupted by noise, the system would mistakenly grasp at noise points in the early stages and then ignore the influence of the noise after a few steps of training. This shows the robustness of our approach in making action decisions.

\begin{figure}[ht]
	\centering
	\includegraphics[width=1\columnwidth]{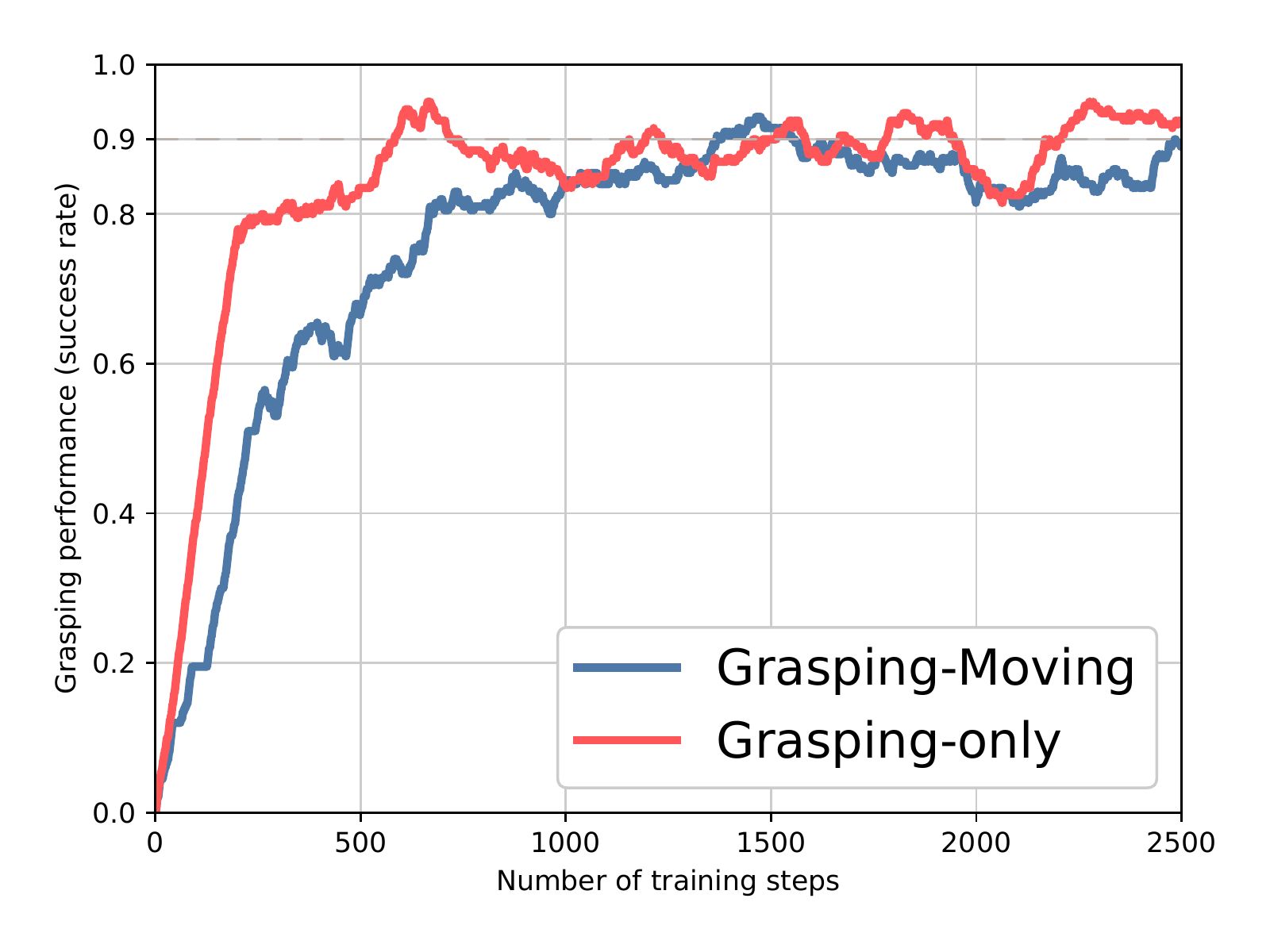}	
	\caption{Evaluating Grasping-only and Grasping-Moving in real-world tests with 20 objects randomly placed.}
	\label{real-baseline}
\end{figure}

\begin{table}[htbp]   
	\caption{\label{t2}Real-world results in the preset environment. }  
	\centering 
	\setlength{\tabcolsep}{2.2mm}{
		\begin{tabular}{l c ccc}    
			\toprule    
			Method    & Completion Rate    & Grasp Success Rate\\
			\midrule
			Grasping-only        & 100$\%$       & 86$\%$  \\
			Grasping-Moving        & 100$\%$       & 92$\%$          \\

			\bottomrule   
	\end{tabular}}  
\end{table}

After that, we compare the performance of the combined grasping-moving strategy versus the grasping-only strategy. This training curve is shown in Fig.~\ref{real-baseline}. As expected, the success rate of grasping in the early stages of training for the grasping-only strategy is better than that of the grasping-moving strategy because it focuses only on grasping. With the increase of the number of training steps, the success rate of the two methods is similar, for random initialization of the environment. Our observations show that in a cluttered environment, grasping-only strategy learns how to fine-tune the grasping position to pick up objects that are hard to grasp, while grasping-moving method tends to move the position of objects and then grasp them. However, as shown in the Table~\ref{t2}, grasping-moving method shows a better grasp success rate in the preset environment (examples of which are shown in the Fig.~\ref{mask}). Notice that grasping-only method can reduce the environmental clutter through failed grasp even if it cannot grasp directly, and then complete the grasping task in a subsequent attempt.

\begin{figure}
	\centering
	\includegraphics[width=1\columnwidth]{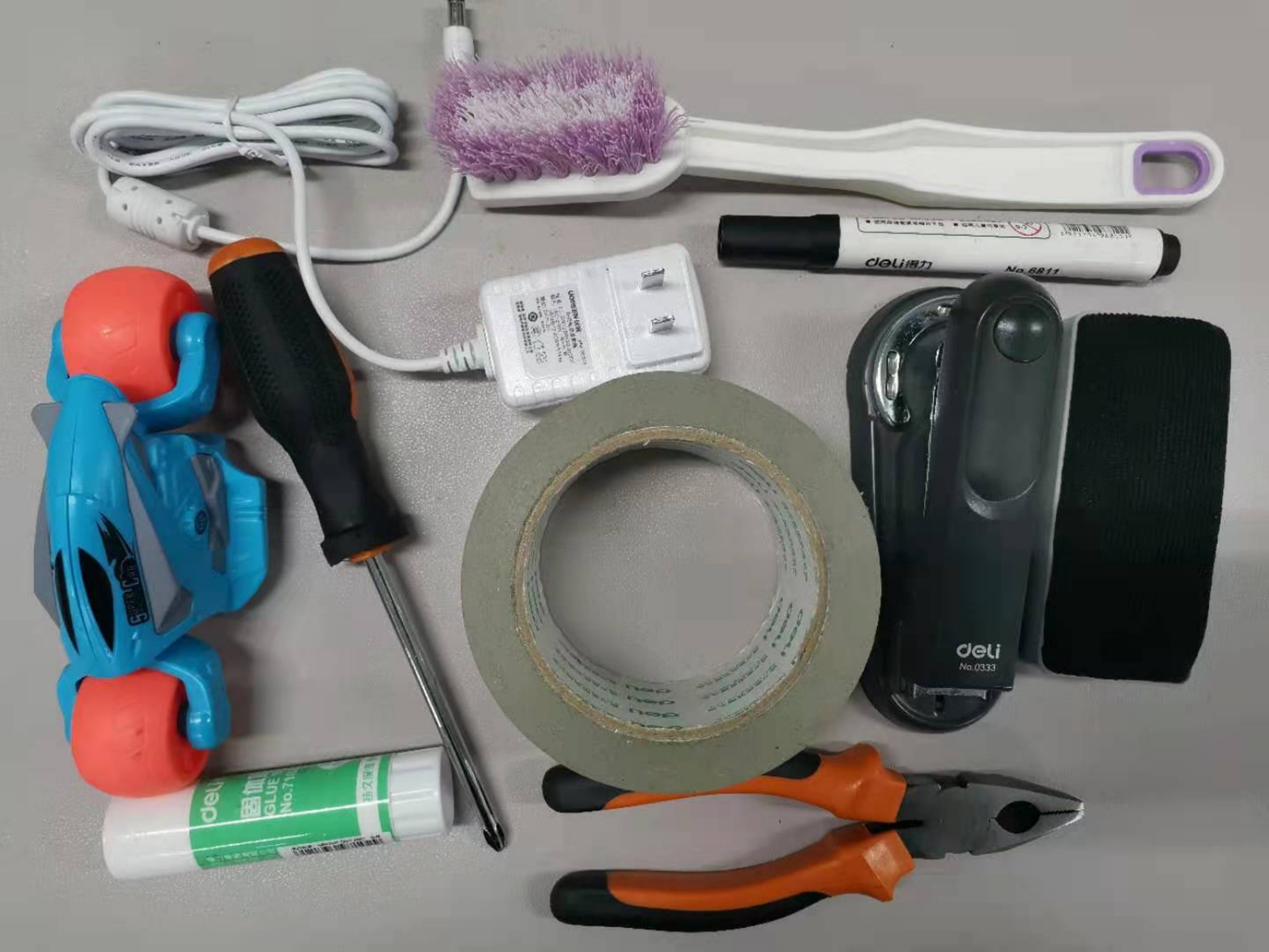}	
	\caption{\textbf{Example of unknown objects} for testing our system’s ability to generalize to novel objects.}
	\label{novel}
\end{figure}

Generalization is a key index for extending the range of applications in industry and logistics automation. We evaluate the ability to generalize to novel objects in a collection of real-world scenes with unseen shapes (examples of which are shown in the Fig.~\ref{novel}). Although our training set only has blocks with regular shape, it is capable of generalizing to the objects with different shapes. As shown in the Table~\ref{t3}, the system achieved an average success rate of~96.6$\%$ for clutter free scenes, and 94.7$\%$ for cluttered scenes. Surprisingly, the system performs well not only on objects similar to the training set such as erasers and glue sticks, but also on objects with extremely irregular shapes. This means that our model has learned a more generalized probability mapping based on model-free reinforcement learning instead of over-fitting to the shapes of the training set. 

\begin{table}[htbp]   
	\caption{\label{t3}Grasp success rate of novel objects. }  
	\centering 
	\setlength{\tabcolsep}{1.2mm}{
		\begin{tabular}{l c ccc}    
			\toprule    
			Object    & Isolation    & Clutter\\
			\midrule
			Charger        & 95$\%$       & 92$\%$  \\
			Stapler        & 96$\%$       & 96$\%$          \\
			Eraser         & 98$\%$      & 97$\%$\\
			Pliers         & 95$\%$      & 95$\%$\\
			Glue stick        & 97$\%$    & 93$\%$     \\
			Marker        & 97$\%$        & 98$\%$         \\
			Brush         & 96$\%$      & 93$\%$\\
			Screwdriver         & 95$\%$    & 93$\%$  \\
			Scotch tape         & 99$\%$      & 97$\%$\\
			Toy         & 98$\%$      & 93$\%$\\
			
			\bottomrule   
	\end{tabular}}  
\end{table}

\section{DISCUSSION AND OUTLOOK}

We have presented a solution for fast learning of grasping and pre-grasping in cluttered environments. We evaluated the performance of the system in both simulation and real world. The model trained in simulation achieved a grasp success rate of 94$\%$ when deployed in real-world cluttered scenarios with minimal real-world data. Since our solution is based on deep reinforcement learning of visual input to learn grasping strategies without object models, it can be extended to a variety of novel objects. The experimental results show that our novel design improves both grasping performance and learning efficiency. As we rely on simulation based learning, we eliminate the need for large amounts of manual data collection and require no human participation in the whole process. Moreover, it takes only 2.5 hours to fine-tune in the real world to achieve a considerably high success rate, which makes our approach feasible for practical deployment. 

We also note the following limitations of our approach. Unlike Song \textit{et al.}~\cite{9126187}, we restrict our action space representation to 4DoF in a discrete space instead of 6DoF in a continuous space. As a result, we limit variety in grasping poses and ignore path planning to reach the grasp point. On a positive note, this facilitates efficient learning. Another limitation is that the robot will persistently complete a grasping action when the controller makes a prediction, even if the environment changes during the execution process. This is likely to lead to failure in some cases, which is also the main factor limiting the success rate of our grasping. We plan to address this in future work by incorporating real-time decision making into the framework.

\addtolength{\textheight}{-8cm}   



%
%

%
%
%
%

\bibliographystyle{IEEEtran}
\bibliography{conf}
\end{document}